\documentclass[lettersize,journal]{IEEEtran}
\usepackage{amsmath,amsfonts}
\usepackage{algorithmic}
\usepackage{algorithm}
\usepackage{makecell}
\usepackage{array}
\usepackage[caption=false,font=normalsize,labelfont=sf,textfont=sf]{subfig}
\usepackage[switch]{lineno}
\usepackage{textcomp}
\usepackage{stfloats}
\usepackage{url}
\usepackage[flushleft]{threeparttable}
\usepackage{bm}
\usepackage{cancel}
\usepackage{verbatim}
\usepackage{graphicx}
\usepackage{cite}
\usepackage[hidelinks]{hyperref}
\newtheorem{definition}{Definition}

\usepackage[some]{background}
\SetBgScale{1}
\SetBgContents{\parbox{20cm}{%
  \Huge Draft:  \today\\[18cm]\rotatebox{180}{\Huge Draft:  \today}}}
\SetBgColor{red}
\SetBgAngle{270}
\SetBgOpacity{0.2}

\begin{document}



\title{CIRO7.2: A Material Network with Circularity of -7.2 and Reinforcement-Learning-Controlled Robotic Disassembler}

\author{Federico Zocco and Monica Malvezzi
\thanks{F. Zocco and M. Malvezzi are with the Department of Information Engineering and Mathematics, University of Siena, Italy. \\ Email: federico.zocco.fz@gmail.com, monica.malvezzi@unisi.it}
\thanks{\emph{(Corresponding author: Federico Zocco)}}
}

\markboth{Journal of \LaTeX\ Class Files,~Vol.~00, No.~0, Month~20XX}%
{Shell \MakeLowercase{\textit{et al.}}: A Sample Article Using IEEEtran.cls for IEEE Journals}


\maketitle

\begin{abstract}
The competition over natural reserves of minerals is expected to increase in part because of the linear-economy paradigm based on take-make-dispose. Simultaneously, the linear economy considers end-of-use products as waste rather than as a resource, which results in large volumes of waste whose management remains an unsolved problem. Since a transition to a circular economy can mitigate these open issues, in this paper we begin by enhancing the notion of circularity based on compartmental dynamical thermodynamics, namely, $\lambda$, and then, we model a thermodynamical material network processing a batch of 2 solid materials of criticality coefficients of 0.1 and 0.95, with a robotic disassembler compartment controlled via reinforcement learning (RL), and processing 2-7 kg of materials. Subsequently, we focused on the design of the robotic disassembler compartment using state-of-the-art RL algorithms and assessing the algorithm performance with respect to $\lambda$ (Fig. \ref{fig:ForTitlePage}). The highest circularity is -2.1 achieved in the case of disassembling 2 parts of 1 kg each, whereas it reduces to -7.2 in the case of disassembling 4 parts of 1 kg each contained inside a chassis of 3 kg. Finally, a sensitivity analysis highlighted that the impact on $\lambda$ of the performance of an RL controller has a positive correlation with the quantity and the criticality of the materials to be disassembled. This work also gives the principles of the emerging research fields indicated as circular intelligence and robotics (CIRO). Source code is publicly available\footnote{\url{https://github.com/ciroresearch/ciro7RL}}.  
\end{abstract}


\section{Introduction} 
\IEEEPARstart{T}he traditional (linear) economy paradigm extracts non-renewable minerals from natural reserves, uses them to manufacture products, sells them to the end user, and eventually disposes of the products at the user discretion. Since this practice is the most common worldwide, it is not surprising that there are two main consequences of increasing concern: the supply uncertainties of critical raw materials used to make green and digital technologies \cite{CRMall,CRM-EU,CRM-US}, and the management of waste \cite{EUwasteData}. The linear economy also fosters competition over natural reserves \cite{competition1} and the stark reality of resources constraints \cite{nygaard2023geopolitical,hageluken2022recycling,lieder2016towards,UoEscarcity,henckens2014metal}. As observed by Henckens \cite{henckens2021scarce}, key factors affecting the access to non-renewable materials are the world population size and the welfare of each individual, with citizens of developed countries currently enjoying the highest consumption rates to satisfy their demands. The scarce raw materials considered by Henckens \cite{henckens2021scarce} are antimony, bismuth, gold, indium, copper along with other eight materials. Enabling future generations to access these materials is at the core of sustainability as defined by the United Nations Brundtland Commission back in 1987 \cite{UN1987sustainability}.  

To mitigate these fundamental issues existing both at national and international levels, the circular economy (CE) paradigm promotes practices such as reuse, reduce, and repair, with the goal of keeping products and materials in use for as long as possible \cite{WhatIsCE}. Since high operational costs are among the major barriers to the implementation of circularity, the cost reductions achievable with robotic and autonomous systems make automation a key enabler of the transition to CE \cite{DigitalizationCosts}. Moreover, robots can avoid the contamination risks resulting from handling waste \cite{zocco2025towards}. 

By definition, the notion of circularity requires to consider the whole material life-cycle, from extraction to disposal, and hence, circularity cannot be measured by merely considering a single process. The holistic perspective is fundamental as pointed out by the Ellen MacArthur Foundation: ``A lack of holistic thinking is one of the reasons industries have been stuck in linear business models up to now; it’s only through rethinking at the system level that we’re likely to get unstuck. [...]'' \cite{CEandHolistic}.   
\begin{figure}
\subfloat[\label{fig:ForTitlePage_a}]{
  \includegraphics[width=0.23\textwidth]{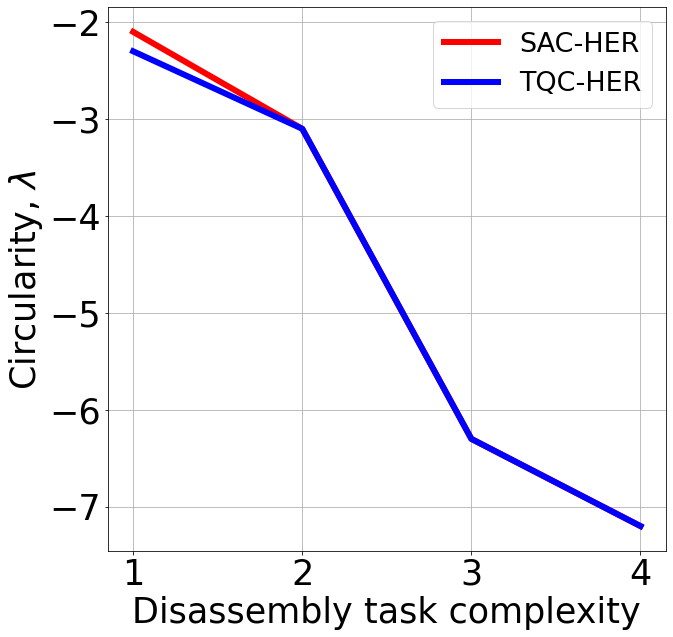}
  }
\subfloat[\label{fig:ForTitlePage_b}]{
  \includegraphics[width=0.24\textwidth]{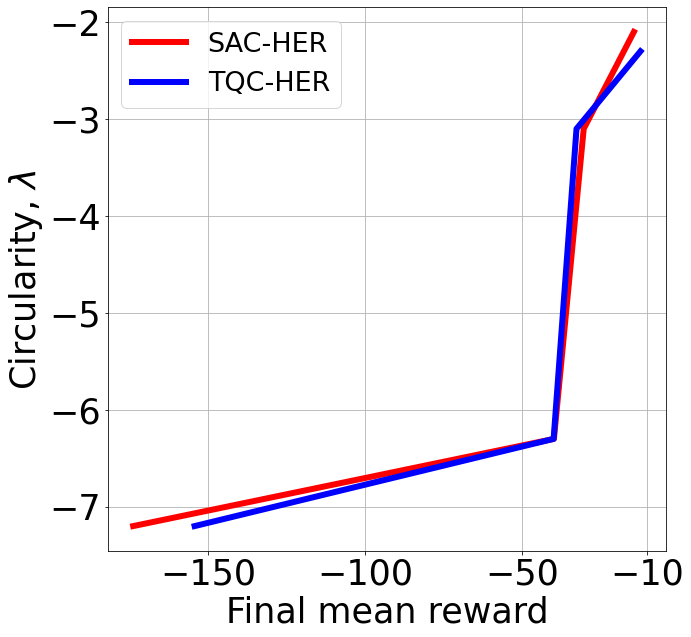}
  }
\caption{(a) Disassembly task complexity vs. $\lambda$. (b) Final mean reward vs. $\lambda$. SAC-HER and TQC-HER are reinforcement-learning algorithms. An higher circularity $\lambda \in (-\infty, 0]$ indicates a more sustainable use of natural resources.}
\label{fig:ForTitlePage}
\end{figure}

In this context, this paper makes the following main contributions:
\begin{itemize}
\item{Designed the first stages of a simulator for reinforcement-learning-controlled robotic disassembly to increase the system circularity (see Section \ref{sub:disassemblerDesign})}.
\item{Enhanced the mathematical formulation of circularity in \cite{zocco2024circular}, which is predicated on compartmental dynamical thermodynamics \cite{haddad2019dynamical}, and hence, has the advantage of being based on fundamental laws of physics in contrast to other circularity metrics \cite{calzolari2022circular}. In addition, the formulation of circularity of this paper can model the material dynamics and be multiform as required in \cite{saidani2024future}, i.e., taking into account reduce, reuse, recycle along with the other circularity practices and also the criticality of the materials (see Section \ref{sub:Foundations}).    
\item{Quantified and analyzed the impact of the performance of the reinforcement-learning (RL) controllers on the system circularity as summarized in Fig. \ref{fig:ForTitlePage} (details in Sections \ref{sub:disassemblerDesign} and \ref{sub:whatIfAnalysis}).}
\item{Overall, we added clarity about the contribution that the robotics community could make for improving circularity, and hence, a sustainable use of natural resources.}
}
\end{itemize}

Throughout the paper, matrices and vectors are indicated with bold upper- and lower-case letters, respectively, while sets are indicated with calligraphic letters. 

The remainder of the paper is organized as follows. Section \ref{sec:RelatedWork} discusses the related work, Section \ref{sec:CompNetDesign} covers the design of the system of which the disassembly stage is part of, and finally Section \ref{sec:Conclusions} gives the conclusions and future work.

\section{Related Work}\label{sec:RelatedWork}
\subsection{Circular Economy and Thermodynamical Material Networks}
A circular economy is based on the so-called ``Rs'', which are practices to be adopted at local scales to improve the material circularity \cite{zorpas2024hidden}. These practices are usually prioritized from the least to the most energy intensive, e.g., \emph{recycling} is one of the last practices in the ranking, whereas \emph{reduce} is one of the first ones \cite{potting2017circular}. A fundamental open question in CE is \emph{where, when, and how to implement the ``Rs'' in order to maximize circularity}, which implies a holistic perspective \cite{abu2024developing}. Addressing this challenging question has motivated the formalization of thermodynamical material networks (TMNs) \cite{zocco2023thermodynamical,zocco2022circularity,zocco2024synchronized}, and subsequently, a notion of circularity predicated on TMNs \cite{zocco2024circular}. 

Material flow analysis (MFA) has become a standard methodology to analyze the circularity of material flows \cite{cullen2022material,brunner2016handbook}. Essentially, MFA is a data-analysis approach based on the principle of mass conservation \cite{brunner2016handbook}. Although TMNs are more mathematically intensive than MFA, they have three main advantages. First, they enhance MFA by adding energy balances to the mass balances; second, they increase the overall modeling and prediction accuracy by leveraging ordinary differential and difference equations derived from thermodynamic principles; and third, they are less data-intensive than MFA because not relying on the availability of data to be analyzed \cite{zocco2023thermodynamical,zocco2025towards}. Hence, TMNs are the approach we adopted in this paper as will be detailed in Section \ref{sec:CompNetDesign}.

\subsection{Robotics for Recovery Chains}
As proved in \cite{zocco2025towards} (specifically, in Proposition 1), the standard form of robot dynamics can be derived from the first law of thermodynamics, and hence, a robot can be considered a thermodynamic compartment of a TMN \cite{zocco2024unification}. While the integration of robots into TMNs is very recent \cite{zocco2025towards}, robots have been proposed in the past to perform recovery-chain tasks such as waste sorting and disassembly. For example, \cite{chen2022robot,koskinopoulou2021robotic,kiyokawa2022challenges} focused on waste sorting, whereas \cite{laili2022optimisation,beghi2023enhancing,LI2018203,qu2023robotic} focused on disassembly. A review of human-robot interaction for disassembly was provided in \cite{yuan2025human}. Although the literature on robotic waste sorting and disassembly is significant, to the best of our knowledge, this is the first paper to measure the robot performance against the circularity of the whole supply-recovery chain (detailed in Section \ref{sec:CompNetDesign} and depicted in Fig. \ref{fig:ForTitlePage}).

\subsection{Reinforcement Learning for Robot Control}
The capability of reinforcement learning systems is progressing fueled by the advances in deep learning. While RL control has shown very promising results in simulated scenarios and games, it still lacks the sampling efficiency and safety guarantees required for deployment in real settings \cite{ju2022transferring}. Ibarz \emph{et al.} \cite{ibarz2021train} encountered other challenges when using RL in real robots such as achieving a reliable and stable learning, designing high-fidelity simulators to train the RL controller, and effectively guiding the exploration phase during learning. Examples of RL uses are \cite{jia2021coach} for snake robot control, \cite{yau2023reinforcement} for diabetes management, \cite{rapetswa2023towards} in cognitive radio networks, and \cite{ghignone2023tc} in autonomous racing. A review of robotic applications is provided in \cite{singh2022reinforcement}.

\section{Compartmental Network Design}\label{sec:CompNetDesign}

\subsection{Foundations}\label{sub:Foundations}
We will begin by giving the definition of a digraph, and then, use it in the definition of a TMN.  
\begin{definition}[\cite{bondy1976graph}]\label{def:Digraph}
A directed graph $D$ or \emph{digraph} is a graph identified by a set of $n_\text{v}$ \emph{nodes} $\{v_1, v_2, \dots, v_{n_\text{v}}\}$ and a set of $n_\text{a}$ \emph{arcs} $\{a_1, a_2, \dots, a_{n_\text{a}}\}$ that connect the nodes. 
\end{definition}
\begin{definition}[\cite{zocco2023thermodynamical}]\label{def:TMN}
A \emph{thermodynamical material network} (TMN) is a set $\mathcal{N}$ of connected thermodynamic compartments, that is, 
\begin{equation}\label{eq:TMNset}
\begin{gathered}
\mathcal{N} = \left\{c^1_{1,1}, \dots, c^{k_\text{v}}_{k_\text{v},k_\text{v}}, \dots, c^{n_\text{v}}_{n_\text{v},n_\text{v}}, \right. \\ 
\left. c^{n_\text{v}+1}_{i_{n_\text{v}+1},j_{n_\text{v}+1}}, \dots, c^{n_\text{v}+k_\text{a}}_{i_{n_\text{v}+k_\text{a}},j_{n_\text{v}+k_\text{a}}}, \dots, c^{n_\text{c}}_{i_{n_\text{c}},j_{n_\text{c}}}\right\}, 
\end{gathered}
\end{equation}
which transport, store, use, and transform a target material, namely, $\beta$. Each compartment is indicated by a \emph{control surface} \cite{moran2010fundamentals} and is modeled using \emph{dynamical systems} derived from a mass balance and/or at least one of the laws of \emph{thermodynamics} \cite{haddad2019dynamical}.
\end{definition}

Specifically, $\mathcal{N} = \mathcal{R} \cup \mathcal{T}$, where $\mathcal{R} \subseteq \mathcal{N}$ is the subset of compartments $c^k_{i,j}$ that \emph{store}, \emph{transform}, or \emph{use} the target material, while $\mathcal{T} \subset \mathcal{N}$ is the subset of compartments $c^k_{i,j}$ that \emph{move} the target material between the compartments belonging to $\mathcal{R} \subseteq \mathcal{N}$. A net $\mathcal{N}$ is associated with its \emph{compartmental diagraph} $M(\mathcal{N})$, which is a digraph whose nodes are the compartments $c^k_{i,j} \in \mathcal{R}$ and whose arcs are the compartments $c^k_{i,j} \in \mathcal{T}$. For node-compartments $c^k_{i,j} \in \mathcal{R}$ it holds that $i = j = k$, whereas for arc-compartments $c^k_{i,j} \in \mathcal{T}$ it holds that $i \neq j$ because an arc moves the material from the node-compartment $c^i_{i,i}$ to the node-compartment $c^j_{j,j}$. The orientation of an arc is given by the direction of the material flow. The superscript $k$ is the identifier of each compartment. The superscripts $k_\text{v}$ and $k_\text{a}$ in (\ref{eq:TMNset}) are the $k$-th node and the $k$-th arc, respectively, while $n_\text{c}$ and $n_\text{v}$ are the total number of compartments and nodes, respectively. Since $n_\text{a}$ is the total number of arcs, it holds that $n_\text{c} = n_\text{v} + n_\text{a}$ \cite{zocco2023thermodynamical,zocco2022circularity,zocco2024synchronized}.

Now, since our goal is to design \emph{circular} networks $\mathcal{N}$ (\ref{eq:TMNset}), we will introduce the notion of circularity after the following definition.
\begin{definition}[\cite{zocco2024circular}]\label{def:ftMassFlow}
A mass or flow is \emph{finite-time sustainable} if either it exits a non-renewable reservoir or it enters a landfill, an incinerator, or the natural environment as a pollutant.
\end{definition}
The locations mentioned in Definition \ref{def:ftMassFlow}, i.e., reservoirs, landfills, incinerators, and the environment, are thermodynamic compartments $c^k_{i,j} \in \mathcal{N}$ as they obey the laws of thermodynamics \cite{haddad2017thermodynamics}. We can now define the circularity of $\mathcal{N}$ (\ref{eq:TMNset}).  
\begin{definition}\label{def:circularity}
The time-window circularity $\lambda(\mathcal{N}; t) \in (-\infty, 0]$ is defined as
\begin{equation}\label{eq:circularity}
\lambda(\mathcal{N}; t) = - \frac{1}{t_\text{f}} \int^{t_\text{f}}_{0} \mu_{\text{f,b}} \overline{m}_{\textup{f},\textup{b}}(t) + \Delta\dot{\overline{m}}_{\textup{f},\textup{c}}(t) \,\, \text{d}t.
\end{equation}
\end{definition}
In (\ref{eq:circularity}), $m_{\text{f},\text{b}}(t)$ is the net finite-time sustainable mass transported in batches (e.g., solids transported on trucks) and $\dot{m}_{\text{f},\text{c}}(t)$ is the net finite-time sustainable flow transported continuously (e.g., fluids transported through pipes). The quantity
\begin{equation}
\overline{m}_{\text{f},\text{b}}(t) = c_{\text{f}, \text{b}} m_{\text{f},\text{b}}(t)
\end{equation}
is the \emph{weighted} net finite-time sustainable mass transported in batches, where $c_{\text{f}, \text{b}} \in (0,1]$ is the criticality coefficient indicating to what extent a material is critical, e.g., $c_{\text{f}, \text{b}} > 0.8$ in the case of critical raw materials \cite{CRM-EU,CRM-US}, while $\mu_{\text{f},\text{b}}$ is the functionality coefficient such that
\begin{equation}\label{eq:mufbDefinition}
\mu_{\text{f},\text{b}} = 1 + l,
\end{equation}
where $l$ is the number of \emph{functional} discarded batches, i.e., batches of products or parts sent to incineration even if they are not faulty. The parameter $\mu_{\text{f},\text{b}}$ in (\ref{eq:circularity}) takes into account the functionality of the discarded products or parts since one of the key principles of circularity is to retain products and materials \emph{at their highest value} \cite{secondPrinCE} (e.g., disposing of a batch of reusable bottles yields $l = 1$, whereas a batch of broken bottles yields $l = 0$; thus, discarding working items  reduces circularity more than the disposal of faulty items).
Similarly, $\dot{\overline{m}}_{\text{f},\text{c}}(t) = c_{\text{f}, \text{c}} m_{\text{f},\text{c}}(t)$ is the \emph{weighted} net finite-time sustainable flow of material transported continuously, where $c_{\text{f}, \text{c}} \in (0,1]$ is the criticality coefficient of the flowing material. Moreover, $\Delta > 0$ is a constant interval of time introduced as a multiplying factor in order to convert the flow $\dot{\overline{m}}_{\text{f},\text{c}}(t)$ into a mass, and hence, to make the sum with $\overline{m}_{\text{f},\text{b}}(t)$ physically consistent. The choice of $\Delta$ is arbitrary, but its value must be kept the same for any calculation of $\lambda(\mathcal{N}; t)$ in order to make comparisons. Finally, $t_\text{f}$ is the final time for the calculation of circularity.

\subsection{Network Design}
\begin{figure}
\includegraphics[width=0.45\textwidth]{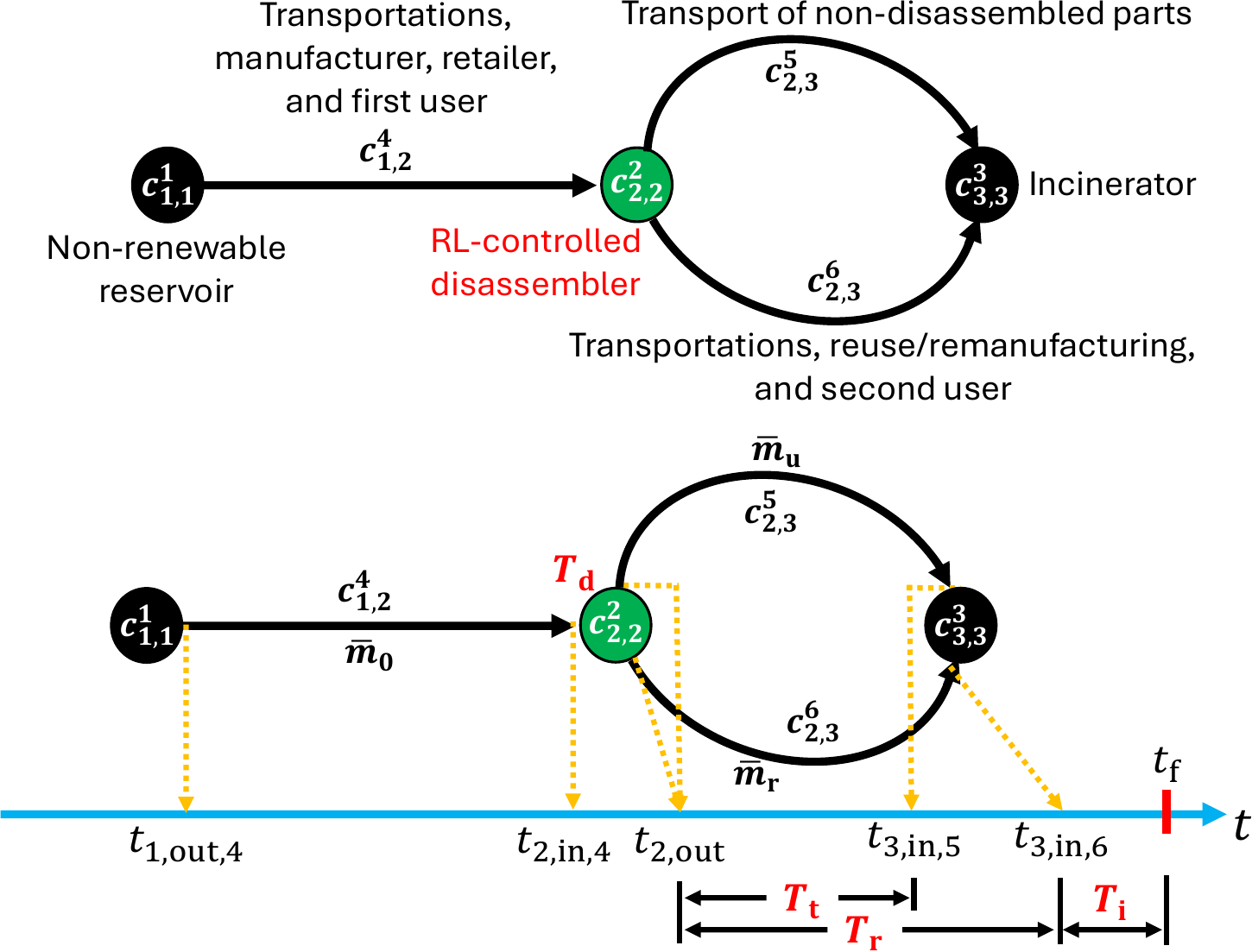}
\centering
\caption{Compartmental diagraph of $\mathcal{N}_\text{s}$ (\ref{eq:Nsolids}). At the top, the description of each compartment; at the bottom, the masses and times are indicated with respect to the temporal axis. In green the disassembler compartment, which is considered in this paper.}
\label{fig:compDiagraphs_solids}
\end{figure}

The material network considered in this paper is
\begin{equation}\label{eq:Nsolids}
\mathcal{N}_\text{s} = \{c^1_{1,1},c^2_{2,2},c^3_{3,3}, c^4_{1,2}, c^5_{2,3}, c^6_{2,3}\}
\end{equation}
and it is shown in Fig. \ref{fig:compDiagraphs_solids}. This network is valid for solid materials and not for fluids because it involves the disassembly stage ($c^2_{2,2}$).
The mass of non-renewable raw materials leaves the reservoir ($c^1_{1,1}$) at $t$ = $t_{1,\text{out},4}$ $\equiv$ 0. Consider that $\mathcal{N}_\text{s}$ processes 2 materials, namely, $\beta_1$ and $\beta_2$. Let $m^0_{\text{f,b},i}$ be the mass of material $\beta_i$ extracted at $t$ = 0, with $i \in \{1,2\}$. Thus, the weighted mass of finite-time sustainable material extracted at $t$ = 0 is   
\begin{equation}
\overline{m}_0 = c_{\text{f,b},1} m^0_{\text{f,b},1} + c_{\text{f,b},2} m^0_{\text{f,b},2},  
\end{equation}
where $c_{\text{f,b},1} \in (0,1]$ and $c_{\text{f,b},2} \in (0,1]$ are the criticality coefficients of the materials $\beta_1$ and $\beta_2$, respectively.

As shown at the top of Fig. \ref{fig:compDiagraphs_solids}, the disassembler ($c^2_{2,2}$) processes the products made of $\beta_1$ and $\beta_2$ after their first use (final stage of $c^4_{1,2}$). The disassembler has two output flows indicated at the bottom of Fig. \ref{fig:compDiagraphs_solids}: one flow is the one in $c^5_{2,3}$ and it involves the mass of parts that are not disassembled, namely, $\overline{m}_\text{u}$. These parts are directly sent to the incinerator ($c^3_{3,3}$). The other flow involves the mass of parts that are disassembled successfully, and then, reused, namely $\overline{m}_\text{r}$. From the principle of mass conservation, it holds that     
\begin{equation}
\overline{m}_0 = \overline{m}_\text{u} + \overline{m}_\text{r}.  
\end{equation}
Let $s \in [0,100]$ be the percentage success measuring to what extent the disassembly stage is successful, it follows that  
\begin{equation}
\overline{m}_{\text{u}} = \overline{m}_0\left(1-\frac{s}{100}\right)
\end{equation}
and
\begin{equation}
\overline{m}_{\text{r}} = \overline{m}_0 \frac{s}{100}.
\end{equation}
Since the mass transfer in $\mathcal{N}_\text{s}$ (\ref{eq:Nsolids}) is a dynamical system, let us now specify the time instants at which the masses enter or exit the compartments. The time instants are indicated at the bottom of Fig. \ref{fig:compDiagraphs_solids}, where $t_{i,\text{out},j}$ is the time at which the material exits compartment $i$ for compartment $j$, while $t_{i,\text{in},j}$ is the time at which the material enters the compartment $i$ from compartment $j$. Specifically, $\overline{m}_0$ exits the non-renewable reservoir at $t_{1,\text{out},4} \equiv 0$. Then, the weighted mass $\overline{m}_0$ enters the disassembly facility ($c^2_{2,2}$) in $t_{2,\text{in},4}$. The disassembly stage lasts $T_\text{d}$. The disassembled and not-disassembled masses of materials (i.e., $\overline{m}_\text{r}$ and $\overline{m}_\text{u}$, respectively) exit the disassembler at $t_{2,\text{out}}$, and hence, $t_{2,\text{out}}$ = $t_{2,\text{in},4}$ + $T_\text{d}$. The material mass $\overline{m}_\text{u}$ enters the incinerator at $t_{3,\text{in},5}$, while the reused material does so at $t_{3,\text{in},6} > t_{3,\text{in},5}$. The incineration lasts $T_\text{i}$ after which the life of the material ends at $t$ = $t_\text{f}$. The time necessary to transport $\overline{m}_\text{u}$ is $T_\text{t}$ = $t_{3,\text{in},5}$ - $t_{2,\text{out}}$, while the time for reusing $\overline{m}_\text{r}$ is $T_\text{r}$ = $t_{3,\text{in},6}$ - $t_{2,\text{out}}$. 

From Definition \ref{def:ftMassFlow}, it follows that the weighted net mass of finite-time sustainable materials transported in batches in $\mathcal{N}_\text{s}$ (i.e., solids) is
\begin{equation} 
\begin{aligned}
\overline{m}_\text{f,b}(t) {}
& = 
\begin{cases}
\overline{m}_0, \quad 0 \leq t < t_{2,\text{in},4} + T_\text{d}+T_\text{t} \\
\overline{m}_0 + \overline{m}_0\left(1 - \frac{s}{100} \right), \quad \parbox[t]{.105\textwidth}{$t_{2,\text{in},4} + T_\text{d}+T_\text{t} \leq t < t_{2,\text{in},4} + T_\text{d} + T_\text{r}$} \\
\overline{m}_0 + \overline{m}_\text{u} + \overline{m}_\text{r}, \quad \parbox[t]{.14\textwidth}{$ t_{2,\text{in},4} + T_\text{d} + T_\text{r} \leq t < t_{2,\text{in},4} + T_\text{d} + T_\text{r} + T_\text{i}$.}
\end{cases}
\end{aligned}
\end{equation}
Regarding the functionality coefficient $\mu_{\text{f,b}}$, we consider that the non-disassembled batch $m_\text{u}$ is sent to the incinerator even if not faulty, whereas the reused batch $m_\text{r}$ has no functionality when it is sent to incineration. Hence, from (\ref{eq:mufbDefinition}), it follows that 
\begin{equation}
l = 1 \Rightarrow \mu_{\text{f,b}} = 2.
\end{equation}

Thus, the dynamics of circularity (\ref{eq:circularity}) in $\mathcal{N}_\text{s}$ is
\begin{equation}
\begin{aligned}\label{eq:lambdaExplicit}
\lambda(\mathcal{N}_{\text{s}}; t, \overline{m}_0, s, T_\text{d}) {} 
& = - \frac{1}{t_\text{f}}\int_{0}^{t_\text{f}} 2\overline{m}_\text{f,b}(t) + \Delta\cancelto{0}{\dot{\overline{m}}_{\text{f,c}}(t)} \quad \text{d}t \\ 
& = - \frac{2}{t_\text{f}}\biggl[ \overline{m}_0\biggl(t_{2,\text{in},4} + T_\text{d}+T_\text{t}\biggr) \\ 
& + \biggl(\overline{m}_0 + \overline{m}_0\biggl(1 - \frac{s}{100} \biggr)\biggr) \biggl(T_\text{r} - \\
& T_\text{t}\biggr)
+ 2\overline{m}_0 T_\text{i}\biggr], 
\end{aligned}
\end{equation}
where
\begin{equation}
t_{\text{f}} = t_{2,\text{in},4} + T_\text{d} + T_\text{r} + T_\text{i}. 
\end{equation}

As visible in (\ref{eq:lambdaExplicit}), the disassembler ($c^2_{2,2}$) affects the circularity $\lambda$ through the percentage success $s$ and the disassembly time $T_\text{d}$. Hence, the next section will cover the design of $c^2_{2,2}$ with the aim of increasing $\lambda$. To extract a numerical value of circularity (\ref{eq:lambdaExplicit}), we considered the values summarized in Table \ref{tab:ValuesOfParam}.  
\begin{table}
\centering
\caption{Values used for the parameters. The variables $s$, $T_d$, $m^0_{f,b,1}$, and $m^0_{f,b,2}$ change as covered in Section \ref{sub:disassemblerDesign}.}
\label{tab:ValuesOfParam}
\begin{tabular}{cc} 
Parameter & Value \\
\hline
$l$ & 1 \\
$c_{\text{f,b},1}$ & 0.1 \\
$c_{\text{f,b},2}$ & 0.95 \\
$t_{1\text{out},4}$ & 0 s\\
$t_{2,\text{in},4}$ & 2,592,000 s (i.e., 1 month)\\
$T_\text{t}$ & 3,600 s (i.e., 1 hour)\\
$T_\text{r}$ & 2,592,000 s (i.e., 1 month)\\
$T_\text{i}$ & 86,400 s (i.e., 1 day)\\
\hline
\end{tabular}
\end{table}

\subsection{Disassembler Compartment Design}\label{sub:disassemblerDesign}
\begin{figure*}
\subfloat[2 parts, 1 target\label{fig:2P1T}]{
  \includegraphics[width=0.24\textwidth]{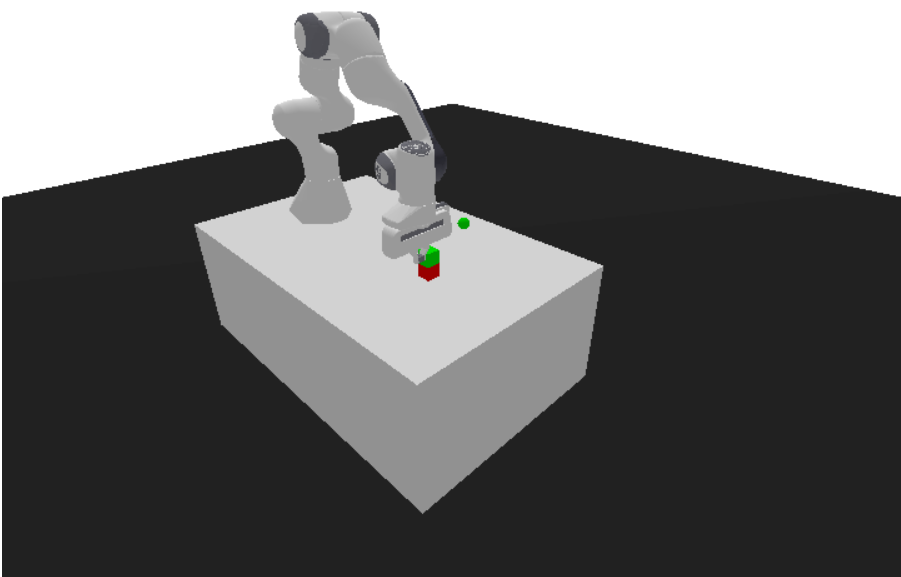}
  }
\subfloat[2 parts, 2 targets\label{fig:2P2T}]{
  \includegraphics[width=0.24\textwidth]{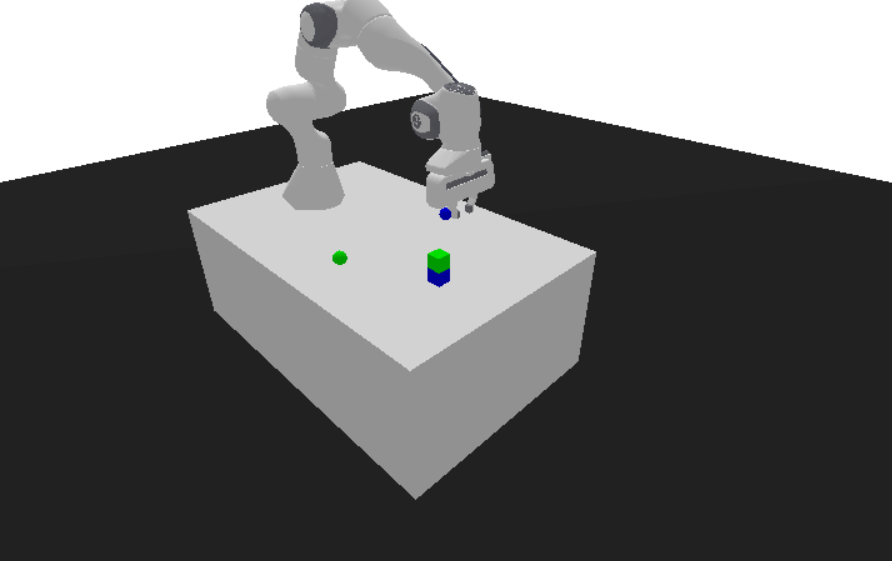}
  }
\subfloat[4 parts, 2 targets, 2 obstacles\label{fig:4P2T2O}]{
  \includegraphics[width=0.24\textwidth]{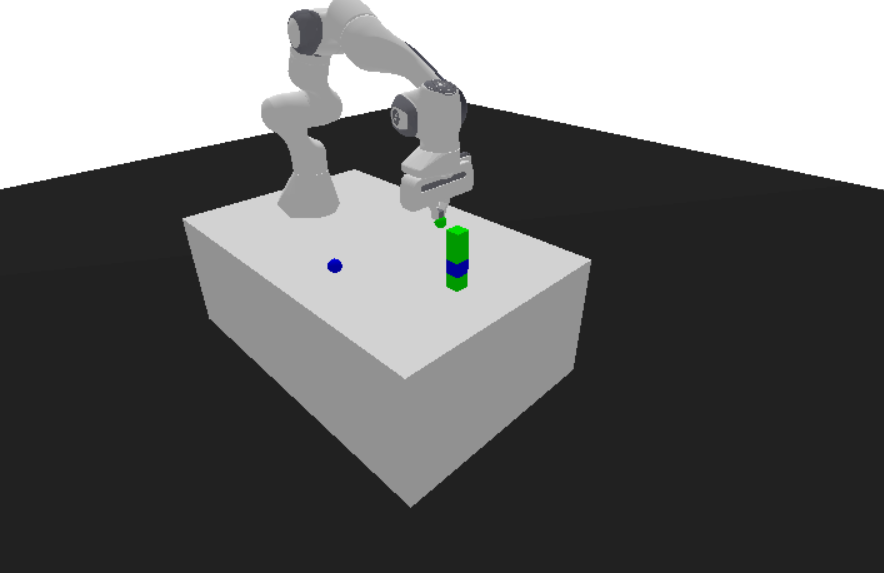}
  }
\subfloat[4 parts, 2 targets, 2 obstacles, 1 chassis\label{fig:4P2T2OC}]{
  \includegraphics[width=0.24\textwidth]{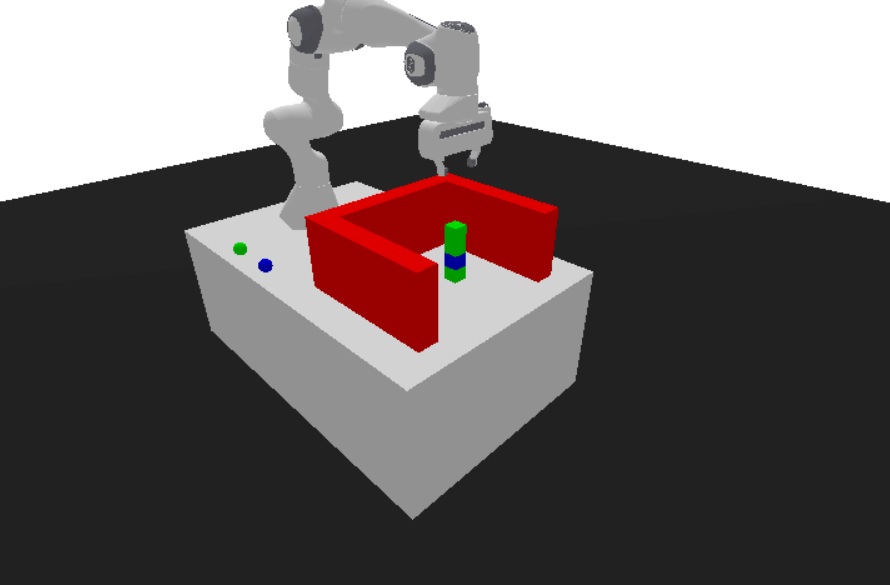}
  }
\caption{Visual simulation of disassembler compartment performance after training ($c^2_{2,2}$ in Fig. \ref{fig:compDiagraphs_solids}).}
\label{fig:ShowEnvironments}
\end{figure*}

We considered four disassembly tasks of increasing complexity to be performed by a Franka Emika Panda robot. Hence, we build on the \emph{panda-gym} library \cite{gallouedec2021panda}. The four tasks (i.e., RL environments) are detailed below and their visual simulations are shown in Fig. \ref{fig:ShowEnvironments}. We compared three RL algorithms from the \emph{Stable-Baselines3} library \cite{raffin2021stable} to train the robot controllers: the soft-actor critic (SAC) \cite{haarnoja2018soft}, the truncated quantile critics (TQC) \cite{kuznetsov2020controlling}, and the twin delayed deep deterministic policy gradient (TD3) \cite{fujimoto2018addressing}. Each learning algorithm was enhanced with hindsight experience replay (HER) \cite{andrychowicz2017hindsight}. 
Over the following, we consider that each part to be disassembled weights 1 kg and is made either of material $\beta_1$ or $\beta_2$.
\begin{figure*}
\subfloat[Pushing of SAC-HER controller  \label{fig:SACpush}]{
  \includegraphics[width=0.24\textwidth]{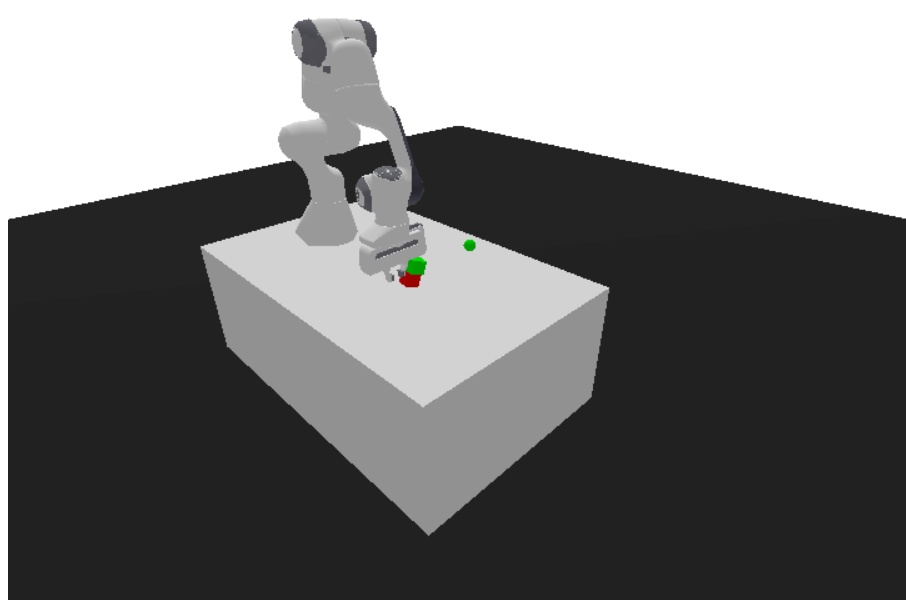}
  }
\subfloat[Picking of TQC-HER controller\label{fig:TQCpick}]{
  \includegraphics[width=0.24\textwidth]{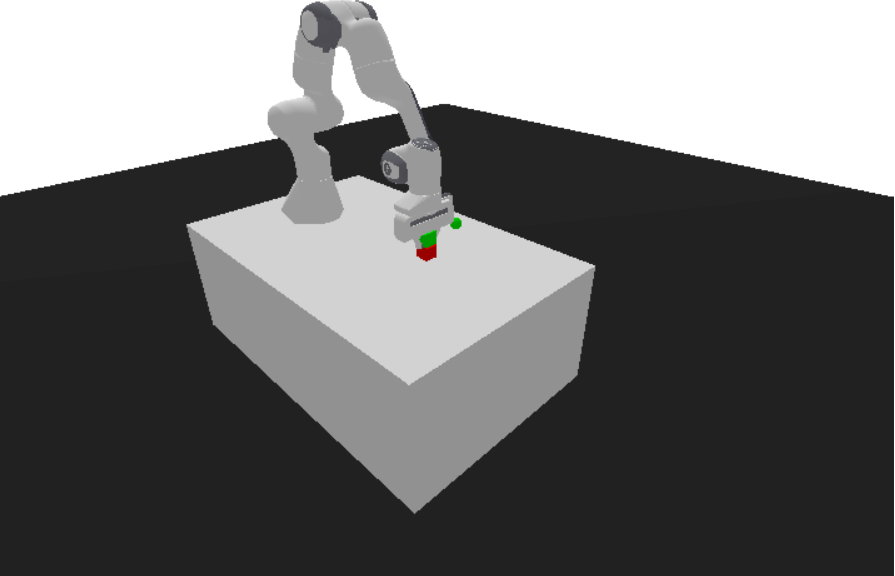}
  }
\subfloat[SAC-HER stops to collide\label{fig:SACstopColliding}]{
  \includegraphics[width=0.24\textwidth]{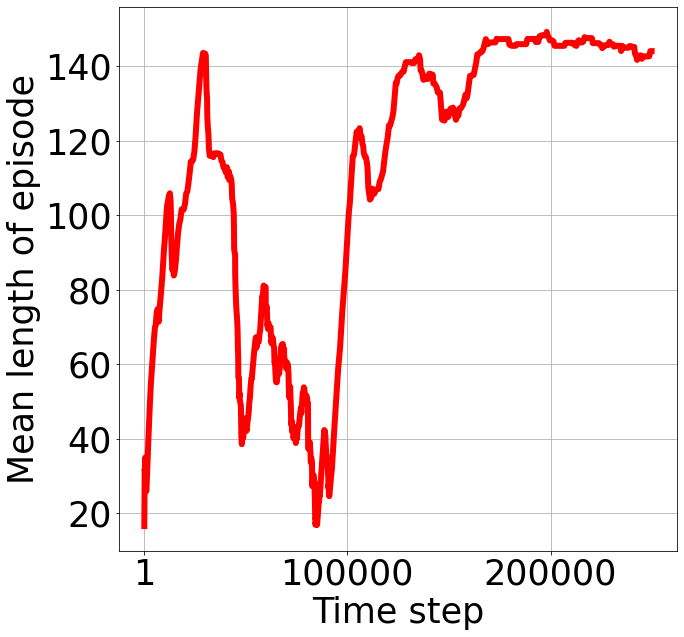}
  }
\subfloat[TQC-HER stops to collide\label{fig:TQCstopColliding}]{
  \includegraphics[width=0.24\textwidth]{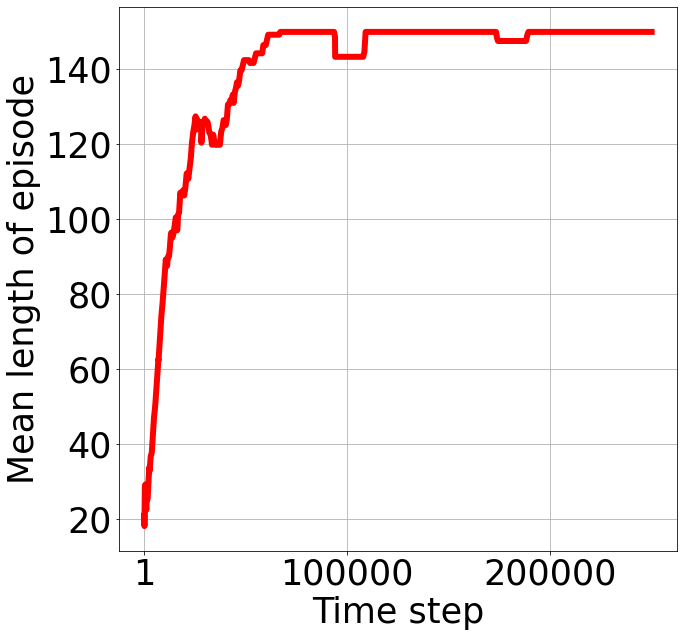}
  }
\caption{Further analysis of the robot behaviors after training (Figs. \ref{fig:SACpush} and \ref{fig:TQCpick}) and during training (Figs. \ref{fig:SACstopColliding} and \ref{fig:TQCstopColliding}). Specifically, Figs. \ref{fig:SACpush} and \ref{fig:TQCpick} regard the scenario with 2 parts and 1 target, while Figs. \ref{fig:SACstopColliding} and \ref{fig:TQCstopColliding} regard the scenario with the chassis.}
\label{fig:AnalysisOfLearnings}
\end{figure*}

\textbf{2 Parts and 1 Target:} The first disassembly task consists of separating two stacked parts modeled as boxes. The robot must pick and place the box at the top in a target location indicated by a sphere in Fig. \ref{fig:2P1T}. One part is made of $\beta_1$ and the other part is made of $\beta_2$, thus $m^0_{\text{f,b},1} = m^0_{\text{f,b},2}$ = 1 kg, and hence, $\overline{m}_0$ = 0.1 + 0.95 = 1.05 kg.
The training settings and performance of the controllers in one training for this task are reported in Table \ref{tab:results-2P1T}, where 
\begin{equation}
\zeta = r_\text{e} - r_{s},
\end{equation}
with $r_\text{e}$ and $r_\text{s}$ denoting the mean reward over 100 episodes at the end and at the start of the training, respectively. The best performing algorithms, namely, SAC and TQC, increase their reward during training and, during testing, yield a percentage success $s$ of 100\% and 80\%, respectively. The disassembly time $T_\text{d}$ is estimated considering that, according to \cite{gallouedec2021panda}, one simulated time step of the RL controller is equivalent to approximately 40 milliseconds of the real time. Equation (\ref{eq:circularity}) yields that the circularity with SAC and TQC is -2.1 and -2.3, respectively, while it reduces to -3.1 with TD3 because it has $s$ = 0\% as a consequence of its learning failure in this single training. Figures \ref{fig:SACpush} and \ref{fig:TQCpick} show that SAC and TQC perform the task in different ways. Specifically, the former pushes the part at the top towards the target (Fig. \ref{fig:SACpush}), whereas the latter picks and drags both parts (Fig. \ref{fig:TQCpick}). As a consequence, $T_\text{d}$ of the SAC controller is lower than that of the TQC controller (0.4 sec vs. 0.8 sec).      
\begin{table*}
\centering
\caption{Performance of training and test of different algorithms in the case of disassembly of 2 parts with 1 target. Trainings were executed on a Colab NVIDIA L4. In bold the resulting circularity (\ref{eq:lambdaExplicit}).}
\label{tab:results-2P1T}
\begin{threeparttable}
\begin{tabular}{ccc|cccc|cccccc} 
 & \thead{Max episode \\ length} & \thead{Time \\ steps} & \thead{Train. time \\ (h:min:sec)} & $r_\text{s}$ & $r_\text{e}$ & $\zeta$ & $(c_{\text{f,b},1}, c_{\text{f,b},2})$ & \thead{$m^0_{\text{f,b},1}$ \\ (kg)} & \thead{$m^0_{\text{f,b},2}$ \\ (kg)} & $T_\text{d}$ (sec) & s & $\lambda$\\ 
\hline
SAC-HER & 50 & 4.5 $\times 10^5$ & 2:41:46 & -50 & -14.3 & 35.7 & (0.1, 0.95) & 1 & 1 & 0.4 & 100.0 & $\bm{-2.1}$\\
TQC-HER & 50 & 4.5 $\times 10^5$ & 2:44:28 & -50 & -12.2 & 37.8 & (0.1, 0.95) & 1 & 1 & 0.8 & 80.0 & $\bm{-2.3}$\\
TD3-HER & 50 & 4.5 $\times 10^5$ & 1:38:58 & -50 & -50 & 0.0 & (0.1, 0.95) & 1 & 1 & ${86400}$\tnote{*} & 0.0 & $\bm{-3.1}$\\
\hline
\end{tabular}
\begin{tablenotes}
  \item[*] When disassembly fails (i.e., when $s$ = 0), $c^2_{2,2}$ works as a waste collection center where material is stored for $T_\text{d}$ = 86400 seconds (i.e., 1 day). 
  \end{tablenotes}
\end{threeparttable}
\end{table*}

\textbf{2 Parts and 2 Targets:} In this disassembly task, the robot is asked to bring one part to a target location and the other part to another target. The situation is shown in Fig. \ref{fig:2P2T}, where the targets are indicated with spheres. Similarly to the previous task, one part is made of $\beta_1$ and the other part is made of $\beta_2$, and hence, $\overline{m}_0$ = 1.05 kg. The training and testing results are reported in Table \ref{tab:results-2P2T}. Although all the algorithms increase their reward during training, the selected number of time steps ($1.5\times 10^5$) is not sufficient to perform successfully the task. Thus, $s$ = 0 for all of them, and hence, $\lambda$ = -3.1.
\begin{table*}
\centering
\caption{Performance of training and test of different algorithms in the case of disassembly of 2 parts with 2 targets. Trainings were executed on a Colab NVIDIA Tesla T4. In bold the resulting circularity (\ref{eq:lambdaExplicit}).}
\label{tab:results-2P2T}
\begin{threeparttable}
\begin{tabular}{ccc|cccc|cccccc} 
 & \thead{Max episode \\ length} & \thead{Time \\ steps} & \thead{Train. time \\ (h:min:sec)} & $r_\text{s}$ & $r_\text{e}$ & $\zeta$ & $(c_{\text{f,b},1}, c_{\text{f,b},2})$ & \thead{$m^0_{\text{f,b},1}$ \\ (kg)} & \thead{$m^0_{\text{f,b},2}$ \\ (kg)} & $T_\text{d}$ (sec) & s & $\lambda$\\ 
\hline
SAC-HER & 100 & $1.5 \times 10^5$ & 1:03:19 & -40.0 & -30.3 & 9.7 & (0.1, 0.95) & 1 & 1 & ${86400}$\tnote{*} & 0.0 & $\bm{-3.1}$ \\
TQC-HER & 100 & $1.5 \times 10^5$ & 1:02:27 & -40.2 & -32.6 & 7.6 & (0.1, 0.95) & 1 & 1 & ${86400}$\tnote{*} & 0.0 & $\bm{-3.1}$ \\
TD3-HER & 100 & $1.5 \times 10^5$ & 0:40:00 & -42.9 & -41.3 & 1.6 & (0.1, 0.95) & 1 & 1 & ${86400}$\tnote{*} & 0.0 & $\bm{-3.1}$ \\
\hline
\end{tabular}
\begin{tablenotes}
  \item[*] When disassembly fails (i.e., when $s$ = 0), $c^2_{2,2}$ works as a waste collection center where material is stored for $T_\text{d}$ = 86400 seconds (i.e., 1 day). 
\end{tablenotes}
\end{threeparttable}
\end{table*}

\textbf{4 Parts, 2 Targets, and 2 Obstacles:} In this disassembly task, there are 4 stacked parts as shown in Fig. \ref{fig:4P2T2O}. The robot is asked to pick the part in blue, which is stacked under 2 green parts. Hence, the 2 parts at the top are obstacles to be moved to a target location (the green sphere). In contrast, the part in blue should be moved to the blue sphere. In this scenario with 4 parts, 2 parts are made of $\beta_1$ and 2 parts are made of $\beta_2$. Hence, 
$m^0_{\text{f,b,}1} = m^0_{\text{f,b,}2}$ = 2 kg, and hence, $\overline{m}_0$ = 2.1 kg. The training and test results of the RL controllers are reported in 
Table \ref{tab:results-4P2T2O}. Although both SAC and TQC increase the reward, the number of training time steps is not sufficient to reach the reward saturation. Thus, all the algorithms yield $s$ = 0 during testing, and hence, $\lambda$ = -6.3. Note that the circularity in this task is lower than in the previous disassembly task (-3.1 vs. -6.3) because now there is more material to be disassembled, and hence, an incomplete disassembly ($s$ = 0) is more detrimental to circularity than in the previous task. In other words, \emph{the impact on circularity of the performance of the RL controller has a positive correlation with the mass and the criticality of the materials to be disassembled.} This finding will be discussed further in Section \ref{sub:whatIfAnalysis}.    

\begin{table*}
\centering
\caption{Performance of training and test of different algorithms in the case of disassembly of 4 parts with 2 targets and 2 obstacles. Trainings were executed on a Colab NVIDIA Tesla T4. In bold the resulting circularity (\ref{eq:lambdaExplicit}).}
\label{tab:results-4P2T2O}
\begin{threeparttable}
\begin{tabular}{ccc|cccc|cccccc} 
 & \thead{Max episode \\ length} & \thead{Time \\ steps} & \thead{Train. time \\ (h:min:sec)} & $r_\text{s}$ & $r_\text{e}$ & $\zeta$ & $(c_{\text{f,b},1}, c_{\text{f,b},2})$ & \thead{$m^0_{\text{f,b},1}$ \\ (kg)} & \thead{$m^0_{\text{f,b},2}$ \\ (kg)} &$T_\text{d}$ (sec) & s & $\lambda$\\ 
\hline
SAC-HER & 100 & $2.0 \times {10}^5$ & 1:21:07 & -50.1 & -39.8 & 10.3 & (0.1, 0.95) & 2 & 2 & ${86400}$\tnote{*} & 0.0 & $\bm{-6.3}$\\
TQC-HER & 100 & $2.0 \times {10}^5$ & 1:26:05 & -89.1 & -39.9 & 49.2 & (0.1, 0.95) & 2 & 2 & ${86400}$\tnote{*} & 0.0 & $\bm{-6.3}$\\
TD3-HER & 100 & $2.0 \times {10}^5$ & 1:04:01 & -58.9 & -51.0 & 7.9 & (0.1, 0.95) & 2 & 2 & ${86400}$\tnote{*} & 0.0 & $\bm{-6.3}$\\
\hline
\end{tabular}
\begin{tablenotes}
  \item[*] When disassembly fails (i.e., when $s$ = 0), $c^2_{2,2}$ works as a waste collection center where material is stored for $T_\text{d}$ = 86400 seconds (i.e., 1 day). 
\end{tablenotes}
\end{threeparttable}
\end{table*}

\textbf{4 Parts, 2 Targets, 2 Obstacles, and a Chassis:} Similarly to the previous disassembly task, in this task there are 4 stacked parts with 2 obstacles. In addition, now there is a chassis indicated in red in Fig. \ref{fig:4P2T2OC}. The robot is asked to not collide with the chassis during disassembly (implemented as a penalty and a truncation of the episode). In this task, 2 parts are made of $\beta_1$ and 2 parts are made of $\beta_2$, while the chassis is made of $\beta_1$ (i.e., the material with low criticality since, usually, a chassis is made of aluminum or plastic). Thus, $m^0_{\text{f,b},1}$ = 2 kg and $m^0_{\text{f,b},2}$ = 5 kg, and hence, $\overline{m}_0$ = 2.4 kg. The training and test performance for this task are reported in Table \ref{tab:results-4P2T2OC}. In the selected training time, none of the algorithms reached a reward that resulted in $s > 0$. However, they learned to not collide with the chassis as visible in Figs. \ref{fig:SACstopColliding} and \ref{fig:TQCstopColliding}. Specifically, the two figures show the mean length of an episode during training for one training with SAC and TQC, respectively. The episode length is short at the beginning, i.e., less than 30 time steps, because of the frequent collisions between the robot and the chassis, which result in frequent episode truncations. Subsequently, the episode length increases and saturates at 150, which is the maximum value chosen before training (Table \ref{tab:results-4P2T2OC}). The saturation to 150 indicates that the episodes are no longer truncated by collisions, but by the episode length limit. 

In this task, the circularity is $\lambda$ = -7.2 with all controllers, which is lower than in all previous disassembly tasks because here $\overline{m}_0$ has the highest value (2.4 kg). This confirms what we observed in the previous task, which is that $s$ = 0 is more detrimental to circularity if the mass of materials to be disassembled is larger.           
\begin{table*}
\centering
\caption{Performance of training and test of different algorithms in the case of disassembly of 4 parts and 1 chassis with 2 targets and 2 obstacles. Trainings were executed on a Colab NVIDIA Tesla T4. In bold the resulting circularity (\ref{eq:lambdaExplicit}).}
\label{tab:results-4P2T2OC}
\begin{threeparttable}
\begin{tabular}{ccc|cccc|cccccc} 
 & \thead{Max episode \\ length} & \thead{Time \\ steps} & \thead{Train. time \\ (h:min:sec)} & $r_\text{s}$ & $r_\text{e}$ & $\zeta$ & $(c_{\text{f,b},1}, c_{\text{f,b},2})$ & \thead{$m^0_{\text{f,b},1}$ \\ (kg)} & \thead{$m^0_{\text{f,b},2}$ \\ (kg)} & $T_\text{d}$ (sec) & s & $\lambda$\\ 
\hline
SAC-HER & 150 & $2.5 \times {10}^5$ & 2:15:07 & -367.9 &  -174.0 & 193.9 & (0.1, 0.95) & 5 & 2 & ${86400}$\tnote{*} & 0.0 & $\bm{-7.2}$\\
TQC-HER & 150 & $2.5 \times {10}^5$ & 2:02:02 & -370.8 & -154.4 & 216.4 & (0.1, 0.95) & 5 & 2 & ${86400}$\tnote{*} & 0.0 & $\bm{-7.2}$\\
TQC-HER & 300 & $2.5 \times {10}^5$ & 2:02:57 & -363.9 & -308.5 & 55.4 & (0.1, 0.95) & 5 & 2 & ${86400}$\tnote{*} & 0.0 & $\bm{-7.2}$\\
\hline
\end{tabular}
\begin{tablenotes}
  \item[*] When disassembly fails (i.e., when $s$ = 0), $c^2_{2,2}$ works as a waste collection center where material is stored for $T_\text{d}$ = 86400 seconds (i.e., 1 day). 
\end{tablenotes}
\end{threeparttable}
\end{table*}

\subsection{Sensitivity of $\lambda$ to RL Performance}\label{sub:whatIfAnalysis} 
In this section, we analyze the variation of $\lambda$ to the performance of the RL controller, i.e., to the variables $s$ and $T_\text{d}$.

In practice, $T_\text{d}, T_\text{i}, T_\text{t} << t_{2,\text{in},4}, T_\text{r}$, and hence, \begin{equation}
t_\text{f} \approx t_{2,\text{in},4} + T_\text{r}
\end{equation}
(consider, for example, the case of car parts, smartphones, desktops, and laptops, that could be used for one or more years before disassembly or disposal). Inserting this in (\ref{eq:circularity}), it yields 
\begin{equation}\label{eq:lambdaApprox1}
\lambda \approx -\frac{2 \overline{m}_0}{t_{2,\text{in},4} + T_\text{r}}\biggl[t_{2,\text{in},4} + \biggl(2 - \frac{s}{100}\biggr)T_\text{r}\biggr].
\end{equation}
From (\ref{eq:lambdaApprox1}) it follows that the variation of $\lambda$ with the disassembly time is negligible since $T_\text{d}$ is of the order of hours at the most, whereas $T_\text{r}$ and $t_{2,\text{in},4}$ are of the order of months or years.

Let 
\begin{equation}
\alpha(s) = \frac{t_{2,\text{in},4}+\biggl(2 - \frac{s}{100}\biggr)T_\text{r}}{t_{2,\text{in},4} + T_\text{r}}, 
\end{equation}
with $\alpha(s) \in [1, 2)$ since $s \in [0,1]$ and 
\begin{equation}
\lim_{T_\text{r} \to \infty} \alpha(0) = 2. 
\end{equation}
Specifically,
\begin{equation}
\alpha(s) = 
\begin{cases}
\frac{t_{2,\text{in},4} + 2T_\text{r}}{t_{2,\text{in},4} + T_\text{r}} &\text{if } s = 0 \text{ (min $s$, max $\alpha$})\\
1 &\text{if } s = 100 \text{ (max $s$, min $\alpha$}),
\end{cases}
\end{equation}
and hence,
\begin{equation}\label{eq:lambdaApprox2}
\lambda \approx 
\begin{cases}
-2\overline{m}_0 \frac{t_{2,\text{in},4} + 2T_\text{r}}{t_{2,\text{in},4} + T_\text{r}} &\text{if } s = 0 \text{ (min $s$, min $\lambda$)} \\
- 2\overline{m}_0 &\text{if } s = 100 \text{ (max $s$, max $\lambda$)}.
\end{cases}
\end{equation}
Equations (\ref{eq:lambdaApprox1}) and (\ref{eq:lambdaApprox2}) show that: first, an increase of disassembly success $s$ increases circularity; second, since $\alpha$ is independent of $\overline{m}_0$, the sensitivity of $\lambda$ to $s$ has a positive correlation with $\overline{m}_0$, which confirms that the importance of a performing RL controller increases when there is an increase of the masses and/or criticality of the materials to be disassembled; third, circularity is approximately independent of time in the case of perfect disassembly (i.e., $s$ = 100); fourth, since $\lambda$ has a positive correlation with $\overline{m}_0$, circularity increases when there is a reduction of extracted masses (i.e., $m^0_{\text{f,b,}i}$) and/or material criticality (i.e., $c_{\text{f,b,}i}$).

\section{Conclusions and Future Work}\label{sec:Conclusions}
This paper developed a network of thermodynamic compartments with RL-controlled robotic disassembly to increase the circularity of the whole system. More complex disassembly tasks are more challenging for RL yielding a lower percentage success, and hence, due to the sensitivity of circularity to the performance of the RL controller, result in lower circularity. In addition, the impact of the performance of the RL-controlled disassembler increases when the criticality and/or the quantity of the materials to be disassembled increases. In summary, there are three main actions that increase circularity: increasing the performance of the robotic disassembler when dealing with multi-component products (action for the disassembly compartment), reducing the extraction of non-renewable materials (action for the manufacturer compartment), and reducing the criticality of the materials used to make products (action for the manufacturer compartment).

At a network level, future work will investigate the system circularity in the case of extractions of multiple batches of material performed at different times. At a compartment level, future work will further develop the disassembly simulator and the RL controller, and then, implement the RL controller on a real robot for disassembly tasks.

\section*{Acknowledgments}
The authors thank Giuseppe Saviano at University of Siena for the advice on the panda-gym library and on RL implementation aspects.

\bibliographystyle{IEEEtran}
\bibliography{References}

\vfill

\end{document}